\documentclass[11pt,a4paper]{article}
\usepackage[hyperref]{emnlp-ijcnlp-2019}
\usepackage{times}
\usepackage{latexsym}
\usepackage{amsmath, bm}
\usepackage{mathtools}
\usepackage{xspace}
\usepackage{chngpage}
\usepackage{amssymb}

\aclfinalcopy 

\title{Integrating Text and Image: Determining Multimodal Document Intent in Instagram Posts}

\author{\textbf{Julia Kruk}\textsuperscript{1}\thanks{\hspace{2mm} Work done while Julia (from Cornell University) and
	Jonah were interns at SRI International.},\hspace{2mm}
	\textbf{Jonah Lubin}\textsuperscript{2$*$},\hspace{2mm}
	\textbf{Karan Sikka}\textsuperscript{1},\hspace{2mm} \textbf{Xiao Lin}\textsuperscript{1}, \\
	\textbf{Dan Jurafsky}\textsuperscript{3},\hspace{2mm} \textbf{Ajay
	Divakaran}\textsuperscript{1}\thanks{\hspace{2mm} Corresponding author, ajay.divakaran@sri.com.},\hspace{2mm} \\
	$*$equal contribution \\
        \textsuperscript{1}SRI International, Princeton, NJ \hspace{2mm}
        \textsuperscript{2}The University of Chicago, Chicago, Illinois\hspace{2mm} \\
\textsuperscript{3}Stanford University, Stanford, CA}

\date{}

\begin{document}
\maketitle

\def\etc{etc.\@\xspace}
\def\etal{et al.\@\xspace}
\def\ie{i.e.\@\xspace}
\def\eg{e.g.\@\xspace}

\def\pd{\partial}
\def\grad{\nabla}
\def\R{\mathbb{R}}
\def\d{\boldsymbol{\delta}}
\def\y{\textbf{y}}
\def\l{\boldsymbol{\ell}}
\def\wrt{w.r.t\onedot}
\def\a{\boldsymbol{\alpha}}
\def\vertspace{0.6em}
\newcommand{\mat}[1]{\bm{#1}}

\begin{abstract}
    Computing author intent from multimodal data like Instagram posts requires
    modeling a complex relationship between text and image.
    For example, a caption might evoke an ironic contrast with the image, so
    neither caption nor image is a mere transcript of the other.
    Instead they combine---via what has been called \textit{meaning multiplication} \newcite{bateman2014text}---to create a new meaning
    that has a more complex relation to the literal meanings of text and image.
	Here we introduce a multimodal dataset of $1299$ Instagram posts
    labeled for three orthogonal taxonomies:
    the authorial intent behind the image-caption pair,
    the contextual relationship between the literal meanings of the image and caption,
    and the semiotic relationship between the signified meanings of the image and caption.
	We build a baseline deep multimodal classifier to validate the taxonomy,
    showing that employing both text and image improves intent detection by
    $9.6\%$ compared to using only the image modality, demonstrating the commonality of non-intersective meaning multiplication.
    The gain with multimodality is greatest when the image and caption diverge semiotically.
    Our dataset offers a new resource for the study of the rich meanings
    that result from pairing text and image. The data is available here
    \url{https://github.com/karansikka1/documentIntent_emnlp19}. 
 
\end{abstract}

\section{Introduction}

Multimodal social platforms such as Instagram
let content creators combine visual and textual modalities.
The resulting widespread use of text+image makes
interpreting author intent in multimodal messages an important task for NLP for document understanding.

\begin{figure}[h]
	\centering
    \includegraphics[trim=40 40 40 60,width=\columnwidth]{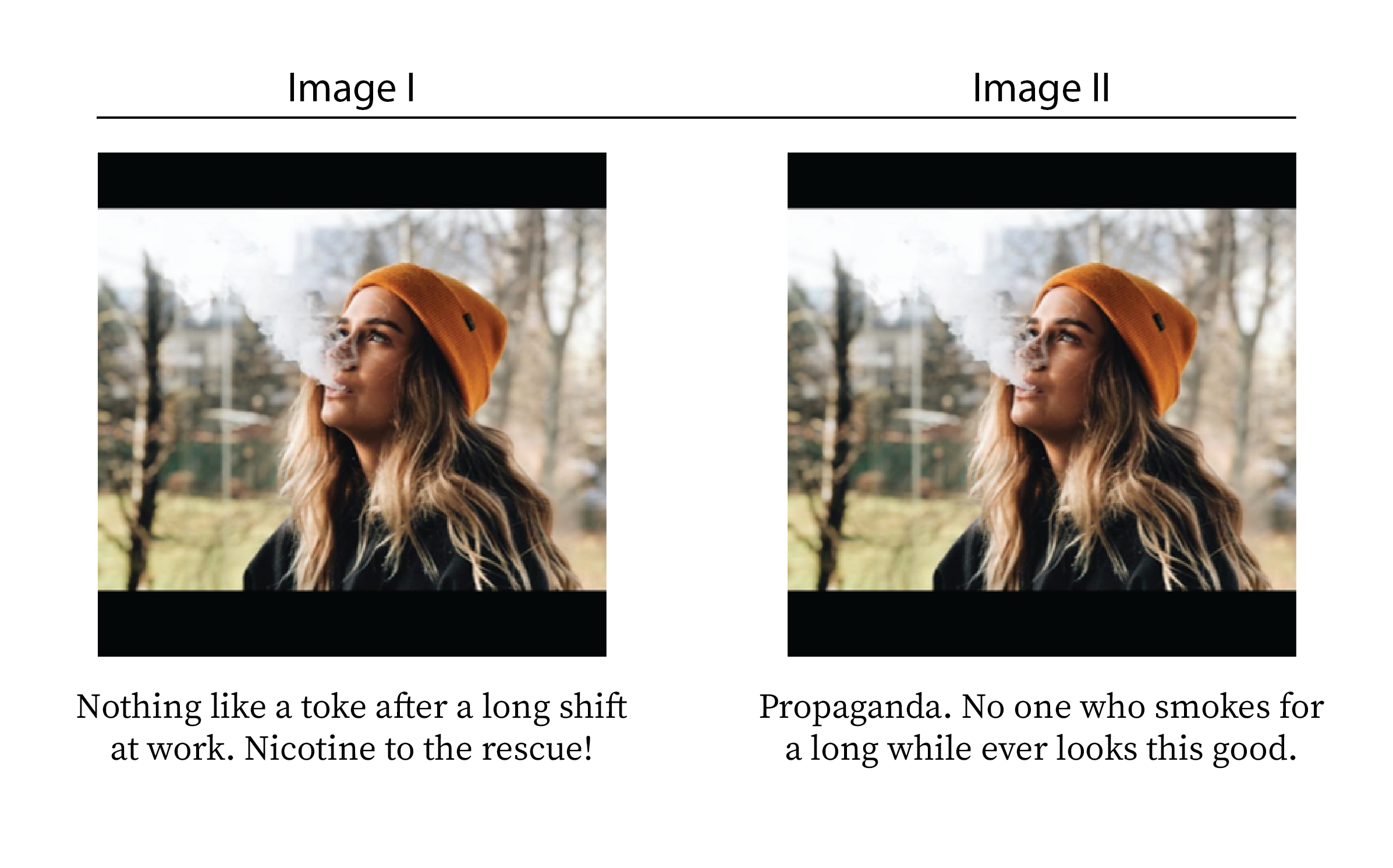}
    \vspace{-1em}
    \caption{Image-Caption meaning multiplication: A change in the caption completely changes the overall meaning of the image-caption pair.}
	\label{fig:1a}
\end{figure}

There are many recent language processing studies of
images accompanied by basic text labels or captions \cite[inter alia]{chen2015microsoft,faghri2018vsepp}.
But prior work on image--text pairs 
has generally been asymmetric, 
regarding either image or text as the primary content, 
and the other as mere complement.
Scholars from semiotics as well as computer science
have pointed out that this is insufficient; often 
text and image are not combined
by a simple  addition or intersection of the component meanings
\cite{bateman2014text, marsh2003taxonomy,zhang2018equal}.

Rather, determining author intent with text+image content requires
a richer kind of meaning composition that has been called
\textit{meaning multiplication} \cite{bateman2014text}:
the creation of new meaning through integrating image and text.
Meaning multiplication
includes simple meaning intersection or concatenation
(a picture of a dog with the label ``dog'', or the label ``Rufus'').
But it also includes more sophisticated kinds of composition,
such as irony  or indirection,
where the text+image integration requires inference that creates a new meaning.
For example in \autoref{fig:1a}, a picture of a young woman smoking 
is given two different hypothetical captions that result in different 
composed meanings. 
In Pairing I, the image and text are parallel, with the
picture used to highlight relaxation through smoking.
Pairing II uses the tension between her image and the implications of her actions to highlight the dangers of smoking. 

Computational models that detect 
complex relationships between text and image 
and how they cue author intent could be significant for many areas,
including the computational study of advertising,
the detection and study of propaganda, 
and our deeper understanding of many other kinds of  persuasive text,
as well as allowing NLP applications to  news media to 
move beyond pure text.

To better understand author intent given such meaning multiplication,
we create three novel taxonomies related to the relationship
between text and image and their combination/multiplication in Instagram posts,
designed by modifying existing taxonomies \cite{bateman2014text, marsh2003taxonomy}  
from semiotics, rhetoric, and media studies.
Our taxonomies measure
the {\bf authorial intent} behind the image-caption pair and two kinds of text-image relations:
the {\bf contextual relationship} between the literal meanings of the image and caption,
and the {\bf semiotic relationship} between the signified meanings of the image and caption.
We then introduce a new dataset, MDID (Multimodal Document Intent Dataset),
with $1299$ Instagram posts covering a variety of topics,
annotated with labels from our three taxonomies.

Finally, we build a deep neural network model for annotating  Instagram posts with the labels
from each taxonomy, and show that
combining text and image leads to better classification,
especially when the caption and the image diverge.  
While our goal here is to establish a computational
framework for investigating multimodal meaning multiplication, 
in other pilot work we
have begun to consider some applications, such as
using intent for social media event detection and for user engagement prediction.
Both these directions highlight the importance
of the intent and semiotic structure of a social media posting in determining its 
influence on the social network as a whole.

\section{Prior Work}

A wide variety of work in multiple fields has explored the relationship between
text and image and extracting meaning, although often
assigning a subordinate
role to either text or images, rather than the symmetric relationship in media such as Instagram posts.
The earliest work in the
Barthesian tradition focuses on advertisements, in which the text serves as merely another connotative aspect to be
incorporated into a larger connotative meaning~\cite{heath1977image}.
\newcite{marsh2003taxonomy} offer a taxonomy of the relationship
between image and text by considering image/illustration pairs found in textbooks or manuals.
We draw on their taxonomy, although as we will see, the connotational aspects of 
Instagram posts require some additions.

For our model of speaker intent, we draw on the classic concept of 
illocutionary acts \cite{austin1962things}
to develop a new taxonomy of illocutionary acts focused on the  kinds of
intentions that tend to occur on social media. For example, 
we rarely see commissive posts on Instagram and Facebook 
because of the focus on information sharing and constructions of self-image.

Computational approaches to multi-modal document understanding have focused on
key problems such as image captioning~\cite{chen2015microsoft,faghri2018vsepp},
visual question answering \cite{goyal2017making,zellers2018recognition,hudson2019},
or extracting the literal or connotative meaning of a post \cite{soleymani2017survey}.
More recent work has explored the 
role of image as context for interaction and pragmatics,
either in dialog \cite{mostafazadeh16,mostafazadeh17},
or as a prompt for users to generate descriptions \cite{bisk19}.
Another important direction has looked at an image's
{\em perlocutionary force} (how it is perceived by its audience), including
aspects such as memorability~\cite{khosla2015understanding},
saliency~\cite{bylinskii2018different}, popularity~\cite{khosla2014makes} and
virality~\cite{deza2015understanding,alameda2017viraliency}.

Some prior work has focused on intention.  \newcite{joo2014visual} and
\newcite{huang2016inferring} study prediction of intent behind politician
portraits in the news.  \newcite{hussain2017automatic} study the
understanding of image and video advertisements, predicting topic, sentiment,
and intent. \newcite{alikhani} introduce a corpus of the coherence
relationships between recipe text and images.  Our work builds on
\newcite{siddiquieICMI2015}, who focused on a single type of intent (detecting
politically persuasive video on the internet) and even more closely on
\newcite{zhang2018equal}, who study visual rhetoric as interaction between the
image and the text slogan in advertisements.  They categorize image-text
relationships  into parallel equivalent (image-text deliver same point at equal
strength), parallel non-equivalent (image-text deliver the same point at
different levels) and non-parallel (text or image alone is insufficient in
point delivery).  They also identify the novel issue of understanding the
complex, non-literal ways in which text and image interacts.
\newcite{WeilandKnowledgeRichImageTextGist} study the non-literal meaning conveyed by image-caption pairs and 
draw on a knowledge-base to generate the gist of the image-caption pair.

\section{Taxonomies}

As \newcite{berger1972} points out in discussing the 
relationship between one image and its caption:
\begin{quotation}
    It is hard to define exactly how the words have changed the image but undoubtedly they have.  (p. 28).
\end{quotation}
We propose three taxonomies
in an attempt to answer Berger's implicit question, two (contextual and semiotic)  to
capture different aspects of the relationship between the image and the caption,
and one to capture speaker intent.
\subsection{Intent Taxonomy}

The proposed intent taxonomy is a generalization and elaboration
of existing rhetorical categories pertaining to illocution, that
targets multimodal social networks like Instagram. 
We developed a set of eight illocutionary intents
from our examination and clustering 
of  a large body of representative
Instagram content, informed by
previous studies of  intent in Instagram
posts.  There is some overlap between categories;
to bound the burden on the annotators, however, we asked
them to identify intent
for the image-caption pairing as a whole and not for the individual components 

\begin{figure}[t]
	\centering
    \includegraphics[trim=40 5 40 20, width=\columnwidth]{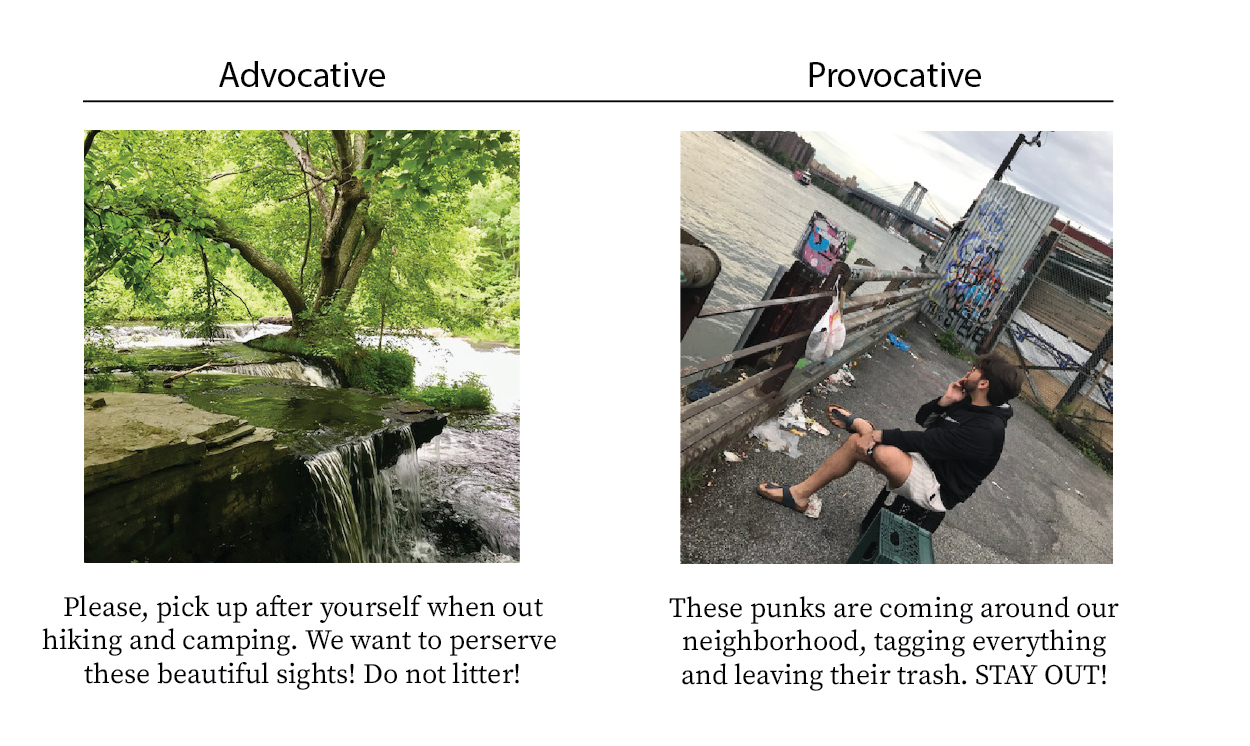}
    \includegraphics[trim=40 5 40 20, width=\columnwidth]{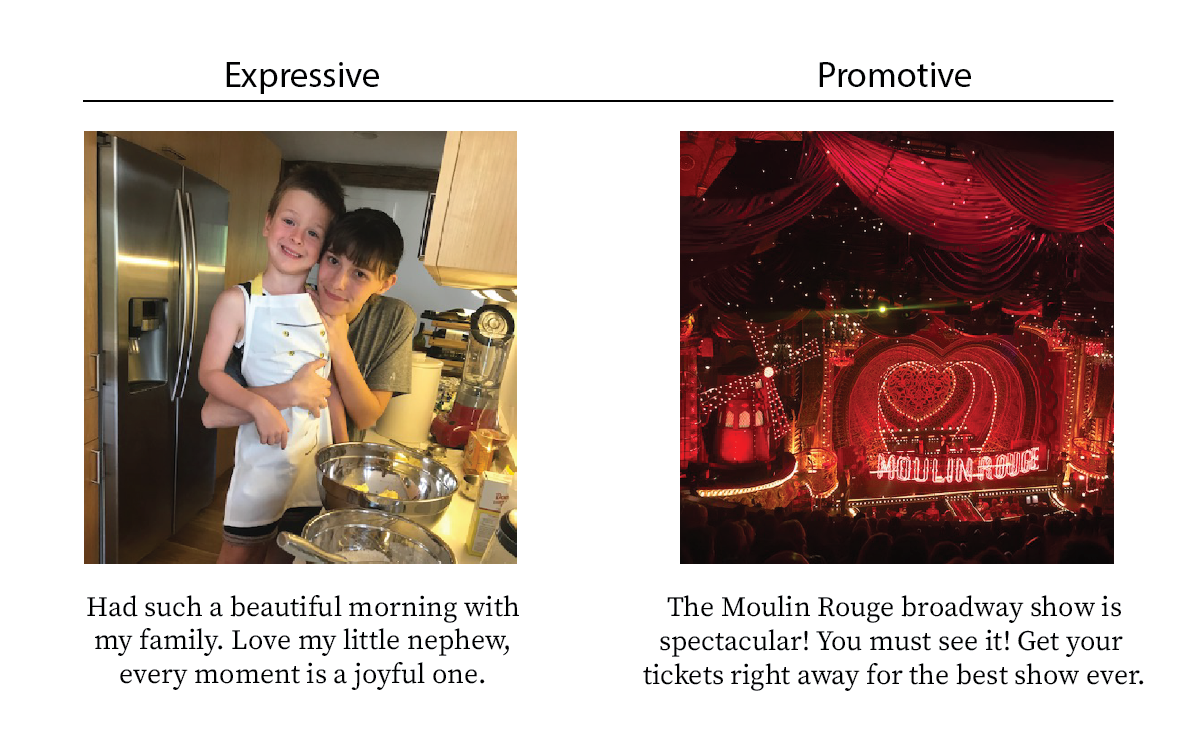}
    \caption{Examples of multimodal document intent: advocative, provocative, expressive and promotive content}
	\label{fig:2}
\end{figure}

For example drawing on 
Goffman's idea of the presentation of self \cite{goffman59},
\newcite{Mahoney16} in their study of Scottish political  Instagram posts
define acts like Presentation of Self, which, following
\newcite{hogan10} we refer to as  {\em exhibition},
or Personal Political Expression, which we generalize to {\em advocative}.
Following are our final eight labels;
Figure~\ref{fig:2} shows some examples.

\begin{enumerate}
    \itemsep 0pt
    \item {\bf advocative}: advocate for a figure, idea, movement, etc. 
    \item {\bf promotive}: promote events, products, organizations etc.
    \item {\bf exhibitionist}: create a self-image reflecting the person, state etc. for the user using selfies, pictures of belongings (e.g. pets, clothes) etc. 
    \item {\bf expressive}: express emotion, attachment, or admiration at an external entity or group.
    \item {\bf informative}: relay information regarding a subject or event using factual language. 
    \item {\bf entertainment}: entertain using art, humor, memes, etc.
    \item {\bf provocative/discrimination}: directly attack an individual or group.
    \item {\bf provocative/controversial}: be shocking. 
\end{enumerate}

\begin{table*}[htb]
	\centering
\begin{tabular}{|cc|}
\hline
	\multicolumn{2}{|c|}{\bf Intent} \\ \hline
    {\em Category} & {\em \# Samples} \\
	Provocative & $84$ \\
	Informative & $119$ \\
	Advocative & $97$ \\
	Entertainment & $310$ \\
	Expositive & $237$ \\
	Expressive & $95$ \\
	Promotive & $162$ \\
\hline
\end{tabular} \hspace{1em}
\begin{tabular}{|cc|}
\hline
	\multicolumn{2}{|c|}{\bf Semiotic} \\ \hline
    {\em Category} & {\em \# Samples} \\
	Divergent & $115$ \\
	Additive & $277$ \\
	Parallel & $712$ \\
\hline
\end{tabular} \hspace{1em}
\begin{tabular}{|cc|}
\hline
	\multicolumn{2}{|c|}{\bf Contextual Relationship} \\ \hline
    {\em Category} & {\em \# Samples} \\
	Minimal & $372$ \\
	Close & $585$ \\
	Transcendent & $147$ \\
\hline
\end{tabular}
    \caption{Counts of different labels in the Multimodal Document Intent Dataset (MDID).}
	\label{table:freq}
\end{table*}
\subsection{The Contextual Taxonomy}
The contextual relationship taxonomy captures  the relationship between the 
literal meanings of the image and text.
We draw on the 
three top-level categories of the \newcite{marsh2003taxonomy} taxonomy,
which distinguished images that are minimally related to the text,
highly related to the text, and related but going beyond it. These three classes---
reflecting Marsh \etal's primary interest in illustration---frame the image only as
subordinate to the text.
We slightly generalize the three top-level categories taxonomy of \newcite{marsh2003taxonomy}
to make them symmetric for the Instagram domain:
\begin{description}
    \itemsep 1pt
    \item [Minimal Relationship:]  The literal meanings of the caption and image overlap very little.
For example, a selfie of a person at a waterfall with the caption ``selfie''. While such a terse caption does nevertheless convey a lot of information, it still leaves out details such as the location, description of the scene, etc. that are found in typical loquacious Instagram captions.
\item [Close Relationship:] The literal meanings of the caption and the image overlap considerably.
For example, a selfie of a person at a crowded waterfall, with the caption ``Selfie at Hemlock falls on a crowded sunny
day''.
\item [Transcendent Relationship:]  The literal meaning of one modality picks up and expands on the literal meaning of
	the other. For example, a selfie of a person at a crowded waterfall with the caption ``Selfie at Hemlock Falls
	on a sunny and crowded day.  Hemlock falls is a popular picnic spot. There are hiking and biking trails, and a
	great restaurant 3 miles down the road ...''.
\end{description}
Note that while the labels ``minimal" and "close" could be thought of as lying on a continuous scale indicating semantic overlap, the label ``transcendent" indicates an expansion of the meaning that cannot be captured by such a continuous scale.

\subsection{The Semiotic Taxonomy}
The contextual taxonomy described above does not deal with the more
complex forms of ``meaning multiplication'' illustrated in \autoref{fig:1a}.
For example, an image of three frolicking puppies with the caption ``My happy family," sends a message of pride in one's pets that is not directly reflected in either modality taken by itself. 
First, it forces the reader to step back and consider what is being signified by the image and the caption,
in effect offering a meta-comment on the text-image relation.
Second, there is a tension between what is signified (a family and a litter of young animals respectively) 
that results  in a richer idiomatic meaning. 

Our third taxonomy therefore captures the relationship between 
what is {\em signified} by the respective modalities, their {\em semiotics}.
We draw on the earlier 3-way distinction of 
\newcite{Kloepfer1977} as modeled by \newcite{bateman2014text}
and the two-way (parallel vs. non-parallel) distinction of \newcite{zhang2018equal}
to classify the semiotic relationship of image/text pairs as divergent, parallel and additive.
A \textit{divergent} relationship occurs when the image and text semiotics pull in opposite directions, creating a gap between the
meanings suggested by the image and text.  A {\em parallel} relationship occurs when the image and text independently contribute to the same meaning. An {\em additive} relationship occurs when the
image and text semiotics  amplify or modify each other.

The semiotic classification is not always homologous to
the contextual. For example, an image of a mother feeding her baby with a caption ``My new small business needs a lot of
tender loving care'', would have a minimal contextual relationship. Yet because both signify loving care and the image intensifies the caption's sentiment,
the semiotic relationship is additive. 
Or a lavish formal farewell scene at an airport with the caption ``Parting is such sweet sorrow'', has
a close contextual relationship because of the high overlap in literal meaning, 
but the semiotics would be additive, not parallel, since the image shows only the leave-taking, while
the caption suggests love (or ironic lack thereof) for the person leaving.  

\begin{figure}[t!]
    \includegraphics[trim=20 0 0 0, width=1.1\columnwidth]{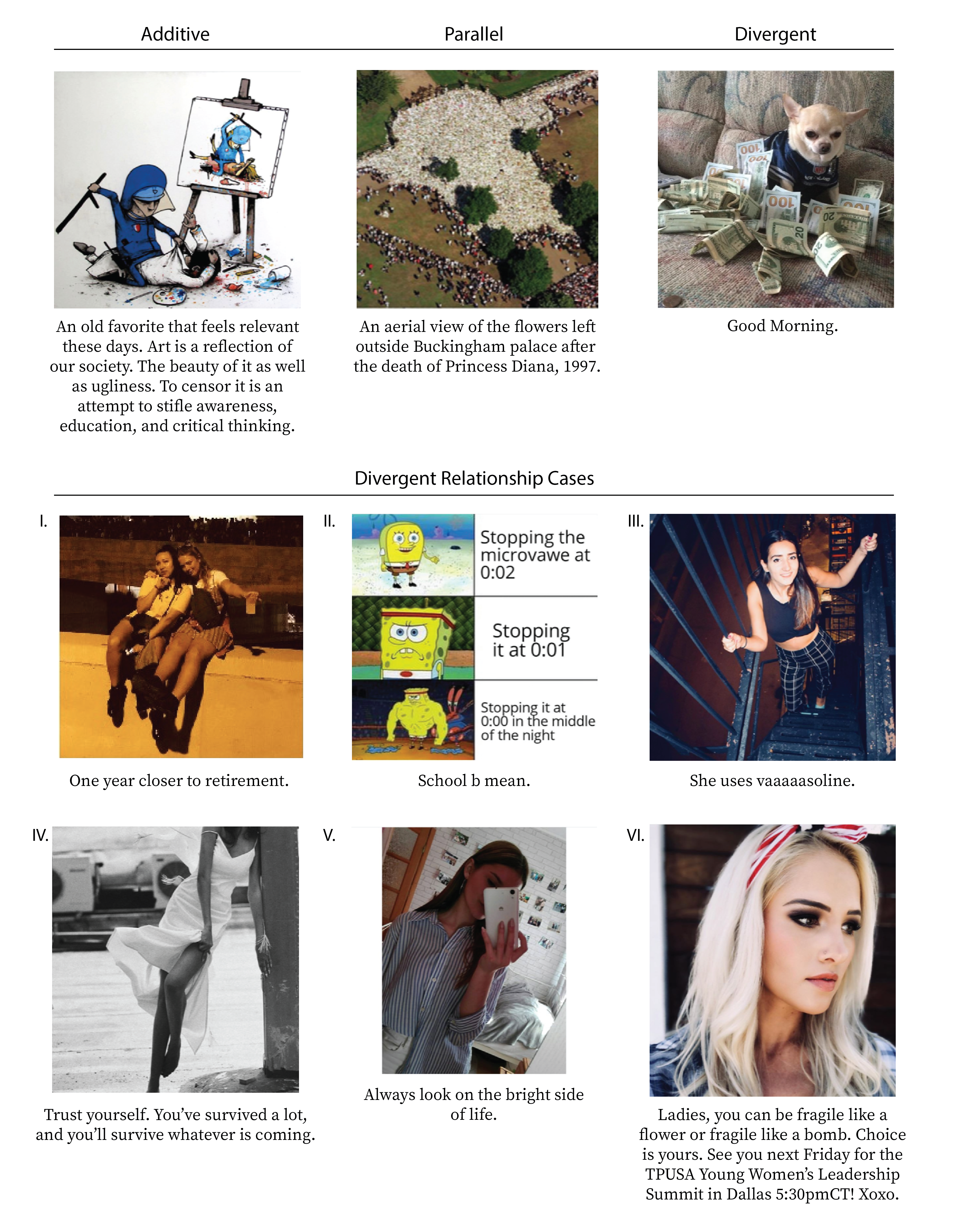}\vspace*{-10pt}
    \caption{The top three images exemplify the semiotic categories. Images I-VI show instances of divergent semiotic relationships.}
	\label{fig:2semiotics}
\end{figure}  
\autoref{fig:2semiotics} further illustrates the proposed semiotic classification. 
The first three image-caption pairs (ICP's) exemplify the three semiotic relationships.
To give further insights into the rich complexity of the divergent category,
the six ICP's below showcase the 
kinds of divergent relationships 
we observed most frequently on Instagram.

ICP I exploits the tension between the reference to retirement expressed in the caption and the youth projected by the two young women in the image to convey irony and thus humor in what is perhaps a birthday greeting or announcement.  Many ironic and humorous posts exhibit divergent semiotic relationships.
ICP II	 has the structure of a classic Instagram meme, where the focus is on the image, and the caption
is completely unrelated to the image. This is also exhibited in the divergent ``Good Morning'' caption in the top row. 
ICP III is an example of a divergent semiotic relationship within an exhibitionist post.
A popular communicative practice on Instagram is to 
combine selfies with a caption that is some sort of inside joke.
The inside joke in ICP III is a lyric from a song a group of friends found funny and discussed the night this photo was taken.
ICP IV is an aesthetic photo of a young woman, paired with a caption that has no semantic elements in common with the photo.
The caption may be a prose excerpt, the author's reflection on what the image made them think or feel,
or perhaps just a pairing of pleasant visual stimulus with pleasant literary material.
This divergent relationship is often found in photography, artistic and other entertainment posts.
ICP V uses one of the most common divergent relationships, in which
exhibitionist visual material is paired with reflections or motivational captions.
ICP V is thus similar to ICP III, but without the inside jokes/hidden meanings common to ICP III.
ICP VI is an exhibitionist post that seems to be common recently among public figures on Instagram.
The image appears to be a classic selfie or often a professionally taken image of the individual, 
but the caption refers to that person’s opinions or agenda(s). This relationship is divergent---there are no common semantic elements in the image and caption---but the pair paints a picture of the individual’s current state or future plans.

\section{The MDID Dataset}
\label{sec:data}

Our dataset, MDID (the \textit{Multimodal Document Intent Dataset}) 
consists of $1299$ public Instagram posts
that we collected  with the goal of developing a rich and diverse
set of posts for each of the eight illocutionary types in our intent taxonomy.
For each intent
we collected at least $16$ hashtags or users likely to yield a high proportion of
posts that could be labeled by that heading.

For the {\em advocative} intent, we selected mostly hashtags advocating and spanning  political or social ideology
such as \#pride and \#maga.
For the {\em promotive} intent we relied on the  \#ad tag that
Instagram has recently begun requiring for sponsored posts.  
For {\em exhibitionist} intent 
we used tags that focused on the self as the most important
aspect of the post  such as
\#selfie and \#ootd (outfit of the day). 
The {\em expressive} posts were retrieved via tags that
actively expressed a stance or an affective intent, such as \#lovehim or \#merrychristmas.
Informative posts were taken from informative
accounts  such as news websites. Entertainment posts drew on an eclectic
group of tags such as \#meme, \#earthporn, \#fatalframes.
Finally, 
provocative posts were extracted via
tags that either expressed a controversial or provocative message or that 
would draw people into being influenced or provoked by the post (\#redpill, \#antifa, \#eattherich, \#snowflake).

\paragraph{Data Labeling:}
Data was pre-processed (for example to convert all albums to single image-caption pairs).
We developed a simple annotation toolkit that displayed an image--caption pair 
and asked the annotator to confirm whether the pair was relevant
(contains both an image and text in English)
and if so
to identify the post's intent
(advocative, promotive, exhibitionist, expressive, informative, entertainment, provocative), 
contextual relationship
(minimal, close, transcendent), and semiotic relationship (divergent, parallel, additive).
Two of the authors collaborated on the labelers manual and then labeled the data by consensus, and any label on which the authors disagreed after discussion was removed.
Dataset statistics are shown in \autoref{table:freq}; see
\url{https://github.com/karansikka1/documentIntent_emnlp19} for the data.

\section{Computational Model}

We train and test a deep convolutional neural network (DCNN) model on the dataset,
both to offer a baseline model for users of the dataset,
and to further explore our hypothesis about meaning multiplication. 

Our model  can take as input either image (Img),  text (Txt) or both (Img + Txt), and consists of modality specific encoders, a
fusion layer, and a class prediction layer.  We use the ResNet-18 network pre-trained on ImageNet as the image
encoder~\cite{he2016deep}.  For encoding captions, we use a standard pipeline that employs a RNN model on word
embeddings.  We experiment with both word2vec type (word token-based) embeddings trained from
scratch~\cite{mikolov2013distributed} and pre-trained character-based contextual embeddings (ELMo)~\cite{peters2018deep}.
For our purpose ELMo character embeddings are more useful since they increase robustness to noisy and often misspelled
Instagram captions.  For the combined model, we implement a simple fusion strategy that first linearly projects encoded
vectors from both the modalities in the same embedding space  and then adds the two vectors. Although naive, this
strategy has been shown to be effective at a variety of tasks such as Visual Question Answering ~\cite{nguyen2018improved}
and image-caption matching~\cite{ahuja2018understanding}. We then use the fused vector  to predict class-wise scores
using a fully connected layer. 

\section{Experiments}

\begin{table*}[!htbp]
   
    \begin{adjustwidth}{-.3in}{-.3in}
	\centering
	\begin{tabular}{c|c|c|c|c|c|c}
		\hline
		Method & \multicolumn{2}{|c|}{Intent} & \multicolumn{2}{|c|}{Semiotic} &
		\multicolumn{2}{|c}{Contextual} \\
		& $ACC$ & $AUC$ & $ACC$ & $AUC$ & $ACC$ & $AUC$  \\
		\hline
		Chance &  $28.1$ & $50.0$ & $64.5$ & $50.0$ & $53.0$ & $50.0$ \\
		Img &  $42.9$ ($\pm 0.0$) & $76.0$ ($\pm 0.5$) & $61.5$ ($\pm 0.0$) & $59.8$ ($\pm 3.0$) &
		$52.5$ ($\pm 0.0$) & $62.5$ ($\pm 1.3$) \\
		Txt-emb &   $42.9$ ($\pm 0.0$) & $72.7$ ($\pm 1.5$) & $58.9$ ($\pm 0.0$)  & $67.8$ ($\pm
		1.7$) & $60.7$ ($\pm 0.5$) & $74.9$ ($\pm 3.0$) \\
		Txt-ELMo &  $52.7$ ($\pm 0.0$) & $82.6$ ($\pm 1.2$) & $61.7$ ($\pm 0.0$)  & $66.5$ ($\pm
		1.9$) & $65.4$ ($\pm 0.0$) & $78.5$ ($\pm 2.1$) \\
		Img + Txt-emb & $48.1$ ($\pm 0.0$) & $80.8$ ($\pm 1.2$) & $60.4$ ($\pm 0.0$)  &
		$69.7$ ($\pm
		1.8$) & $60.8$ ($\pm 0.0$) & $76.0$ ($\pm 2.5$) \\
		Img + Txt-ELMo & $56.7$ ($\pm 0.0$) & $85.6$ ($\pm 1.3$) & $61.8$ ($\pm 0.0$)  & $67.8$ ($\pm
		1.8$) & $63.6$ ($\pm 0.5$) & $79.0$ ($\pm 1.4$) \\
		\hline

	\end{tabular} 
	\end{adjustwidth}
	 \caption{Table showing results with various DCNN models-- image-only (Img), text-only (Txt-emb and Txt-ELMo), and
	combined model (Img + Txt-emb
	and Img + Txt-ELMo). Here \textit{emb} is the model using standard word (token) based embeddings, while
	\textit{ELMo} is
	the pre-trained ELMo based word embeddings \cite{peters2018deep}. The numbers in Table2 are standard deviations across 5 folds. }
	\label{tab:perf1a}
\end{table*}

We evaluate our models on  predicting intent, semiotic
relationships, and image-text relationships from Instagram posts,  using
image only, text only, and  both modalities.     

\subsection{Dataset, Evaluation and Implementation}

We use the 1299-sample MDID dataset (\autoref{sec:data}).
We only use corresponding image and text information for each post and do not use other meta-data
 to preserve the focus on image-caption joint meaning.   We perform basic pre-processing on the captions 
such as removing stopwords and non-alphanumeric characters. We do not perform any pre-processing for images.

Due to the small  dataset, we perform 5-fold
cross-validation for our experiments reporting average performance across all splits. We report  classification accuracy 
(ACC) and also area under the ROC curve (AUC) (since AUC is more robust to class-skew), using macro-average across all classes \cite{jeni2013facing,stager2006dealing}. 

We use a pre-trained ResNet-18 model as the image encoder. For word token based embeddings we use $300$ dimensional vectors
trained from scratch. For ELMo we use a publicly available API
\footnote{\url{https://github.com/allenai/allennlp/blob/master/tutorials/how\_to/elmo.md}}
and use a pre-trained
model with two layers resulting in a $2048$ dimensional input. We use a bi-directional GRU as the RNN model with $256$ dimensional hidden
layers. We set the dimensionality of the common embedding space in the fusion layer to $128$. In case there is a single
modality, the fusion layer only projects features from that modality. We train with the Adam optimizer  with
a learning rate of $0.00005$, which is decayed by $0.1$ after every $15$ epochs. We report results with the best model
selected based on performance on a mini validation set.  

\subsection{Quantitative Results}

We show results in \autoref{tab:perf1a}.  For the intent taxonomy
images are more informative than (word2vec) text
($76\%$ for Img vs $72.7\%$ for Txt-emb)
but with ELMo text  outperforms just using images
($82.6\%$ for Txt-ELMo, $76.0\%$ for Img). 
ELMo similarly improves performance on the contextual taxonomy but not the semiotic taxonomy.

For the semiotic taxonomy, ELMo and word2vec  embeddings perform similarly,
($67.8\%$ for Txt-emb vs. $66.5\%$ for Txt-ELMo),
suggesting that individual words are sufficient for the semiotic labeling task,
and the presence of the sentence context (as in ELMo) is not needed.

Combining visual and textual modalities helps across the board.
For example, for intent taxonomy the joint model Img + Txt-ELMo achieves $85.6\%$ compared to
$82.6\%$ for Txt-ELMo. Images seem to help even more
when using a word embedding based text model
($80.8\%$ for Img + Txt-emb vs. $72.7\%$ for Txt-emb).   
Joint models also improve over single-modality on labeling the image-text relationship and the
semiotic taxonomy.     We show class-wise
performances with the single- and multi-modality models in \autoref{tab:class}. 
It is particularly interesting that in the semiotic taxonomy, multimodality helps the most
with divergent semiotics (gain of $4.4\%$ compared to the image-only model).

\begin{table*}[htbp!]
	\centering
	\scalebox{0.95}{
	\begin{tabular}{|c|c|c|c|}
		\hline
		\multicolumn{4}{|c|}{Intent} \\
		\hline
		Class & Img & Txt-ELMo & Img+ \\
		& & & Txt-ELMo \\ \hline 
		Provocative & $85.5$ & $84.1$ & $90.0$ \\ \hline
		Informative & $77.0$ & $93.9$ & $92.8$ \\ \hline
		Advocative & $84.8$ & $82.4$ & $87.4$  \\ \hline
		Entertainment & $69.0$  & $78.6$ & $80.5$ \\ \hline
		Exhibitionist & $81.7$  & $78.7$  & $84.9$ \\ \hline
		Expressive & $57.9$  & $72.0$  & $73.2$  \\ \hline
		Promotive & $76.3$  & $88.5$  & $90.1$ \\ \hline
		Mean & $76.0$  & $82.6$  & $85.6$ \\ \hline
\end{tabular}} \qquad 
	\scalebox{0.95}{
	\begin{tabular}{|c|c|c|c|}
		\hline
		\multicolumn{4}{|c|}{Semiotic} \\
		\hline
		Class & Img & Txt-emb & Img+ \\
		& & & Txt-ELMo \\ \hline 
		Divergent & $69.8$ & $72.7$ & $77.1$ \\ \hline
		Additive & $55.0$ & $66.7$ & $68.2$ \\ \hline
		Parallel & $54.5$ & $64.3$ & $64.0$  \\ \hline
		Mean & $59.8$  & $67.8$  & $69.7$ \\ \hline
        \multicolumn{4}{c}{}\\
		\hline
		\multicolumn{4}{|c|}{Contextual} \\
		\hline
		Class & Img & Txt-ELMo & Img+ \\
		& & & Txt-ELMo \\ \hline 
		Minimal & $60.9$ & $79.7$ & $81.3$ \\ \hline
		Close & $60.5$ & $73.8$ & $74.6$ \\ \hline
		Transcendent & $66.1$ & $82.0$ & $81.2$  \\ \hline
		Mean & $62.5$  & $78.5$  & $79.0$ \\ \hline
\end{tabular}} 	
	\caption{Class-wise results (AUC) for the three taxonomies with different DCNN models on the MDID dataset. Except for
    the semiotic taxonomy we used ELMo text representations (based on the performance in
    \autoref{tab:perf1a}). 
	}
	\label{tab:class}
\end{table*}

\begin{figure}[h]
	\centering
    \includegraphics[width=0.8\columnwidth]{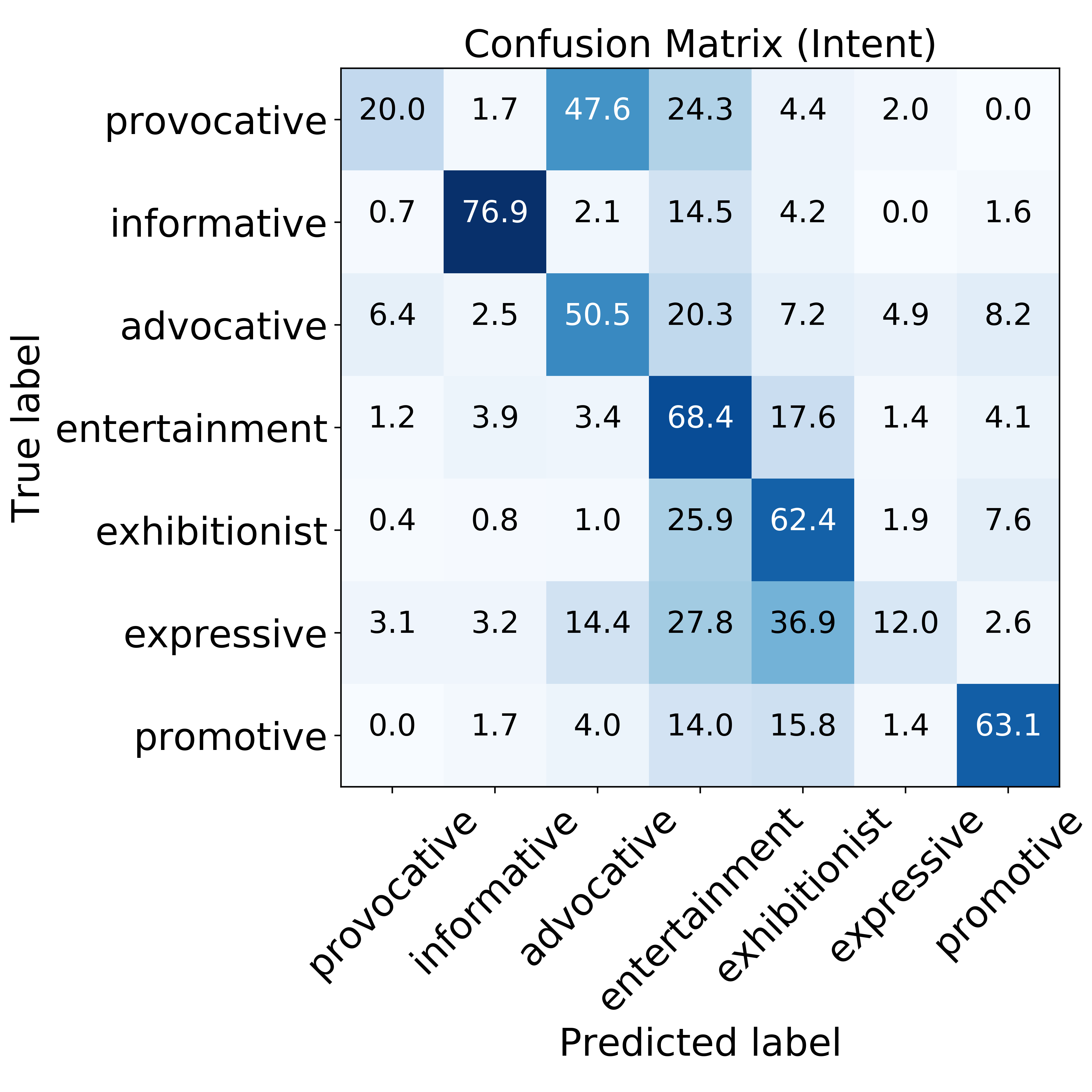}\vspace*{-5pt}
\caption{Confusion between intent classes for the intent classification task. The confusion matrix
	was obtained using the Img + Txt-ELMo model and the results are averaged over the $5$ splits. }
	\label{fig:cm}
\end{figure}

\subsection{Discussion}

In general, using both text and image is helpful,
a fact that is unsurprising since combinations of text and image are known to increase performance
on tasks such as predicting post popularity or user personality \cite{hessel17,wendlandt2017multimodal}.
Most telling, however, were the differences in this helpfulness across items.
In the semiotic category
the greatest gain came when the text-image semiotics were ``divergent''.
By contrast, multimodal models help less when the image and text are additive,
and help the least when the image and text are parallel and provide
less novel information.
Similarly, with contextual relationships, 
multimodal analysis helps the most with
the ``minimal'' category (1.6\%).
This further supports the idea that on social media such as Instagram,
the relation between image and text can be richly divergent and thereby form new meanings. 

The category confusion matrix in \autoref{fig:cm} provides further insights. 
The least confused category is informative. Informative posts are  least similar to the rest of Instagram, since they consist of detached, objective posts, with little  use of first person
pronouns like ``I'' or ``me."  
Promotive posts are  also relatively easy to detect, since they are 
formally informative, telling the viewer the advantages
and logistics of an item or event, with the addition of a persuasive
intent reflecting the poster's personal opinions
(``I love this watch''.).
We found that the {\em entertainment} label is often  misapplied;
perhaps to some extent all posts have a goal of entertaining,
and any analysis must account for this filter of entertainment.
The {\em Exhibitionist} intent tends to be predicted well, likely due to its visual and textual signifiers of individuality (\eg
selfies are almost always exhibitionist, as are captions like ``I love my new hair'').
There is a great deal of confusion, however, between the expressive and exhibitionist categories, since the only distinction lies in whether the post is about a general topic or about the poster,
and between the provocative and advocative categories, perhaps
because both often seek to prove points in a similar way.

With the contextual and semiotic taxonomies, 
some good results are obtained with text alone. In the ``transcendent"
contextual case, it is not necessarily surprising that using text
alone enables 82\% AUC because whenever a caption is really long
or has many adverbs, adjectives or abstract concepts, it is
highly likely to be transcendent. In the ``divergent" semiotics case,
we were surprised that text alone would predict divergence with 72.7\%
AUC.  Examining these cases showed that many of them 
had lexical cues suggesting irony or sarcasm,
allowing the system to infer that
the image will diverge in keeping with the irony. 
There is, however, a consistent improvement when both modalities are
used for both  taxonomies.

\begin{figure*}[htb]
	\centering
   \includegraphics[trim=40 30 40 20,width=1\textwidth]{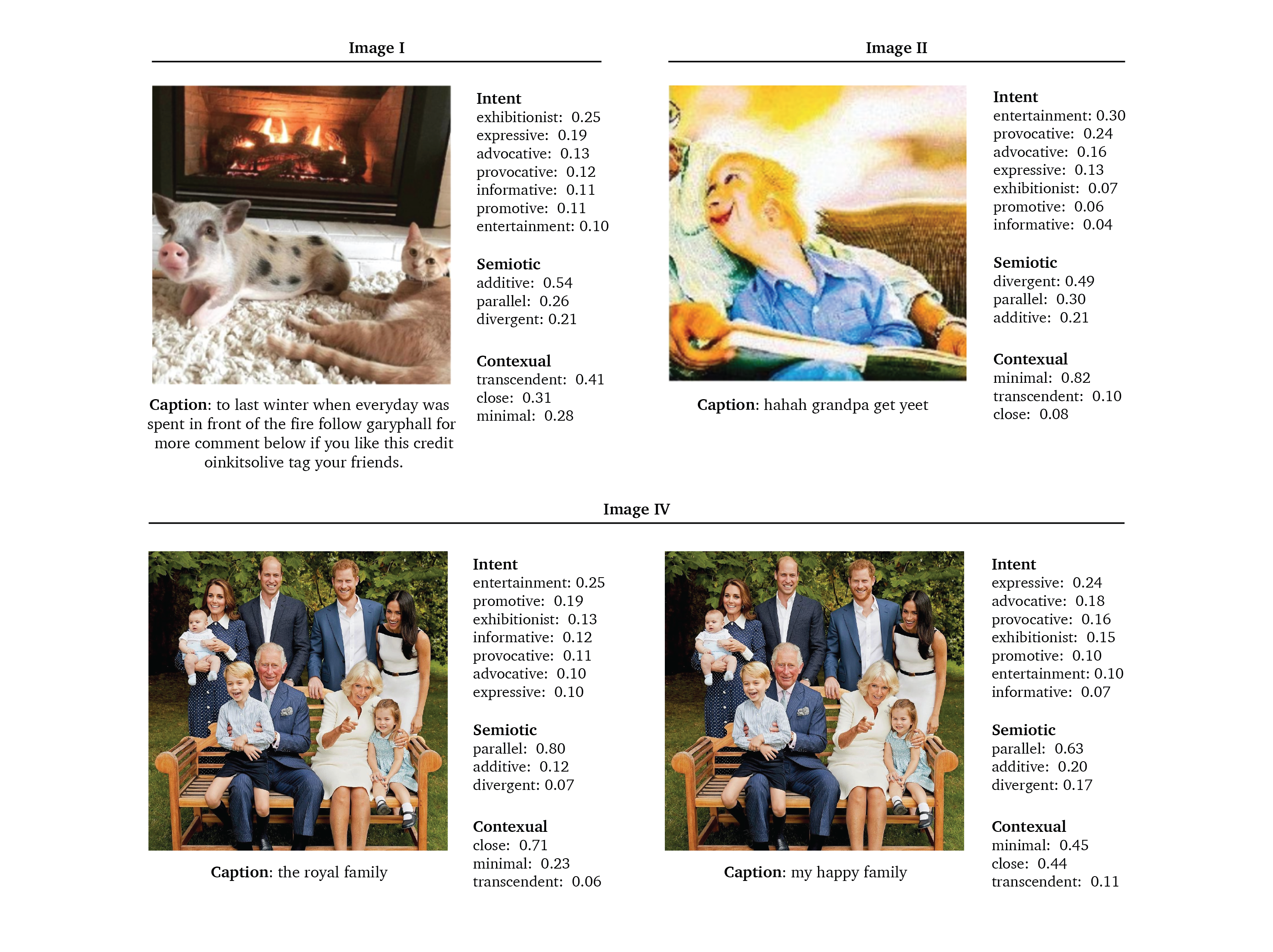}
	\caption{Sample successful output predictions for the three
	taxonomies, showing ranked classes and predicted probabilities.
	In images IV the same image when paired with a different
	caption gives rise to a different intent.} \label{fig:qual_1}
\end{figure*}

\subsection{Sample Outputs}
	
We show some sample successful outputs of the (multimodal) model in \autoref{fig:qual_1},
in which the highest probability class in each of the three dimension corresponds to our gold labels.
The top-left Image-caption pair (Image I) is classified as exhibitionist,
closely followed by expressive;
it is a picture of someone's home with a caption describing a domestic experience.
The semiotic relationship is classified as additive;
the image and caption   together  signify the concept of spending winter at home with pets before the fireplace. The contextual relationship is classified as transcendental;  the caption indeed goes well beyond the image.

The top-right image-caption pair (Image II) is classified as entertainment;  the image caption pair works as an ironic reference to dancing (``yeet'') grandparents,  who are actually reading, in language used usually by young people that a typical grandparent would never use. The semiotic relationship is classified as divergent and the contextual relationship is classified as minimal;  
there is semantic and semiotic divergence of the image-caption pair 
caused by the juxtaposition of youthful references with older people.

To further understand the role of meaning multiplication, we consider the change in intent and semiotic relationships when the same image of the British Royal Family is
matched with two different captions in the bottom row  of \autoref{fig:qual_1} (Image IV).
In both cases the semiotic relationship is parallel, perhaps due to
the match between the multi-figure portrait setting and the word {\em family}.
But the other two dimensions show differences.
When the caption is ``the royal family'' our system classifies
the intent as entertainment; presumably
such pictures and caption pairs often appear on Instagram intending to entertain.
But when the caption is ``my happy family'' the intent is classified as
expressive, perhaps due to the family pride expressed in    
the caption.

\section{Conclusion}

We have proposed a model
to capture the complex \textit{meaning multiplication} relationship between image and text in multimodal Instagram posts.  Our three new taxonomies,
adapted from the media and semiotic literature, allow
the literal, semiotic, and illocutionary relationship between text and image to be coded.
Of course
our new dataset and the baseline classifier models  are just a preliminary
effort, and future work will need to examine larger datasets, consider additional data such as hashtags,
richer classification schemes, and more sophisticated classifiers.
Some of these may be domain-specific.
For example, \newcite{alikhani} show how to 
develop rich coherence relations that model
the contextual relationship between recipe text and accompanying images
(specific versions of Elaboration or Exemplification such as ``Shows a tool used in the step but not mentioned in the text'').
Expanding our taxonomies with richer sets like these is an important goal.
Nonetheless,  the fact that we found multimodal classification to be
most helpful in cases where the image and text diverged 
semiotically points out the importance of these complex relations,
and
our taxonomies, dataset, and tools
should provide impetus for the community to further develop more complex models
of this important relationship.

\section{Acknowledgment}

This project is sponsored by the Office of Naval Research (ONR) under the contract number N00014-17C-1008 and by DARPA
Communicating with Computers (CwC) program under ARO prime contract number W911NF- 15-1-0462. Disclaimer: The views,
opinions, and/or findings expressed are those of the author(s) and should not be interpreted as representing the
official views or policies of the Department of Defense or the U.S. Government. We would like to thank the reviewers for their incisive critique. We would like to thank Prof. Joyce Chai. Ajay Divakaran would like to dedicate
this paper to his mother Mrs. Bharathi Divakaran.

\bibliographystyle{acl_natbib}
\bibliography{biblio}

\end{document}